\documentclass[11pt]{article}
\usepackage[preprint]{acl}

\usepackage{times}
\usepackage{latexsym}
\usepackage[T1]{fontenc}
\usepackage[utf8]{inputenc}
\usepackage{microtype}
\usepackage{inconsolata}
\usepackage{graphicx}
\usepackage{amsmath}
\usepackage{amssymb}
\usepackage{booktabs}
\usepackage{url}

\title{MEDR: Query-Independent Frame Selection via Multi-Signal Event Modeling and Dynamic Rescoring}

\author{
Xinlei Pu\textsuperscript{1}, Weijie Shi\textsuperscript{2}, Wen Yang\textsuperscript{1},
Yi Cao\textsuperscript{1}, Yao Zhao\textsuperscript{3} \\
\bfseries Hao Chen\textsuperscript{4}, Yuanjun Liu\textsuperscript{1}, Wenwei Ding\textsuperscript{5},
Jia Zhu\textsuperscript{6}, Jiajie Xu\textsuperscript{1} \\[0.5ex]
\textsuperscript{1} School of Computer Science and Technology, Soochow University \\
\textsuperscript{2} The Hong Kong University of Science and Technology \\
\textsuperscript{3} Alibaba Group \quad \textsuperscript{4} FiT, Tencent \\
\textsuperscript{5} Suzhou Rural Commercial Bank \\
\textsuperscript{6} School of Education, Zhejiang Normal University
}

\hypersetup{
  pdftitle={MEDR: Query-Independent Frame Selection via Multi-Signal Event Modeling and Dynamic Rescoring},
  pdfauthor={Xinlei Pu, Weijie Shi, Wen Yang, Yi Cao, Yao Zhao, Hao Chen, Yuanjun Liu, Wenwei Ding, Jia Zhu, Jiajie Xu}
}

\begin{document}

\maketitle
\begin{abstract}

Frame selection is a fundamental component of multimodal large language models, enabling long videos to be processed under limited visual-token and computational budgets. Uniform sampling preserves temporal coverage but may miss informative content that appears only briefly. To alleviate this limitation, query-dependent methods can retrieve question-relevant frames. However, because the selected frames depend on the current question, the same visual input cannot be directly shared across different questions, and frame selection must be repeated in multi-turn video dialogue. This motivates us to seek a query-independent frame selection method that preserves the reusability of a fixed visual input while improving the coverage of informative events beyond uniform sampling. We propose \textbf{M}ulti-Signal \textbf{E}vent Modeling and \textbf{D}ynamic \textbf{R}escoring (MEDR), a training-free and query-independent frame selection method. Multi-Signal Event Modeling organizes complementary visual, motion, and text signals into signal-specific temporal events. Dynamic Rescoring then iteratively reevaluates each candidate relative to the current selected set, updating its score according to frame-level signal strength, additional event coverage, and balanced temporal coverage. The resulting fixed frame set is constructed without observing the query and can be reused across different questions. Experiments on Video-MME and LongVideoBench demonstrate that MEDR achieves competitive performance across two video MLLMs. Under the 64-frame budget, MEDR improves Video-MME accuracy over the strongest evaluated query-independent baseline by 1.33--1.78 percentage points. We further evaluate the reuse of a fixed frame set in multi-question video dialogue.

\end{abstract}

\section{Introduction}

Multimodal large language models (MLLMs) extend image-based architectures
to video understanding \cite{lin2024videollava,li2024mvbench,jin2024chatunivi}.
Long videos contain more frames than these models can process efficiently:
additional frames increase visual encoding, prefill, memory, and attention
costs. Frame selection is therefore essential under limited computational
budgets, particularly for long-video benchmarks such as Video-MME,
LongVideoBench, MLVU, and EgoSchema
\cite{fu2025videomme,wu2024longvideobench,zhou2025mlvu,mangalam2023egoschema}.

Uniform sampling preserves temporal coverage but may select similar
content repeatedly while missing brief informative moments. Query-dependent
methods retrieve question-relevant frames, requiring new selections as
the question changes. We study query-independent selection: a fixed frame
set is constructed before any query and reused across questions. Our goal
is to improve the coverage of  informative content while preserving this reusable
visual input.

Achieving this goal presents two main challenges. First, visual changes, motion bursts, and on-screen
text follow complementary temporal patterns that a single global score
does not distinguish. Second, a frame's value changes as the selected set
grows: evidence becomes less useful once a related event or nearby temporal
region is represented. A fixed ranking can therefore concentrate selections
on similar content or short intervals.

We propose MEDR, a training-free and query-independent method with two
modules. Multi-Signal Event Modeling identifies visual, motion, and text
intervals and their representative frames. Dynamic Rescoring updates each
remaining candidate's utility after every selection, combining frame-level
signal strength, additional event coverage, and balanced temporal coverage.
The resulting frame set can be reused across questions without reselection.

Our contributions are summarized as follows:
\begin{itemize}
    \item We propose MEDR, a training-free and query-independent frame
    selection method that constructs a fixed visual input before any
    question is observed. It preserves cross-question reusability while
    improving the coverage of sparsely distributed video content under a
    limited frame budget.

    \item We introduce Multi-Signal Event Modeling to capture complementary
    visual, motion, and text events, together with Dynamic Rescoring to
    update candidate scores according to the evolving selected set.

    \item Experiments on Video-MME and LongVideoBench demonstrate that MEDR
    improves 64-frame Video-MME accuracy over Uniform and MaxInfo
    across both backbones and achieves competitive LongVideoBench
    accuracy. We also evaluate a fixed frame set across different
    questions without query-specific reselection.
\end{itemize}

\section{Related Work}
\label{sec:related-work}

\subsection{Video MLLMs and Training-Free Video Processing}

Video MLLMs connect sampled-frame features to language models through
projection, pooling, or learned bridges. Video-LLaVA, VideoChat2, and
Chat-UniVi extend image understanding to temporal inputs
\cite{lin2024videollava,li2024mvbench,jin2024chatunivi}.
Long-video systems use memory, timestamps, keyframes, or adaptive compression
\cite{song2024moviechat,ren2024timechat,tan2024koala,shen2025longvu}.
VideoAgent and VideoTree actively localize evidence conditioned on the question
\cite{fan2024videoagent,wang2025videotree}.

Training-free approaches reorganize pretrained-model inputs: FreeVA uses
multiple frames, IG-VLM uses image grids, SlowFast-LLaVA combines temporal
pathways, and TS-LLaVA mixes thumbnails with sampled tokens
\cite{wu2024freeva,kim2024igvlm,xu2024slowfastllava,qu2024tsllava}.
Keyframe--token selection and token pruning further reduce inference cost
\cite{song2026ktv,chen2024fastv,zhang2025sparsevlm}.
MEDR constructs the input frame set under a fixed budget while leaving the
MLLM and its visual-token processing unchanged.

\label{sec:method}

\begin{figure*}[t]
    \centering
    \includegraphics[width=\textwidth]{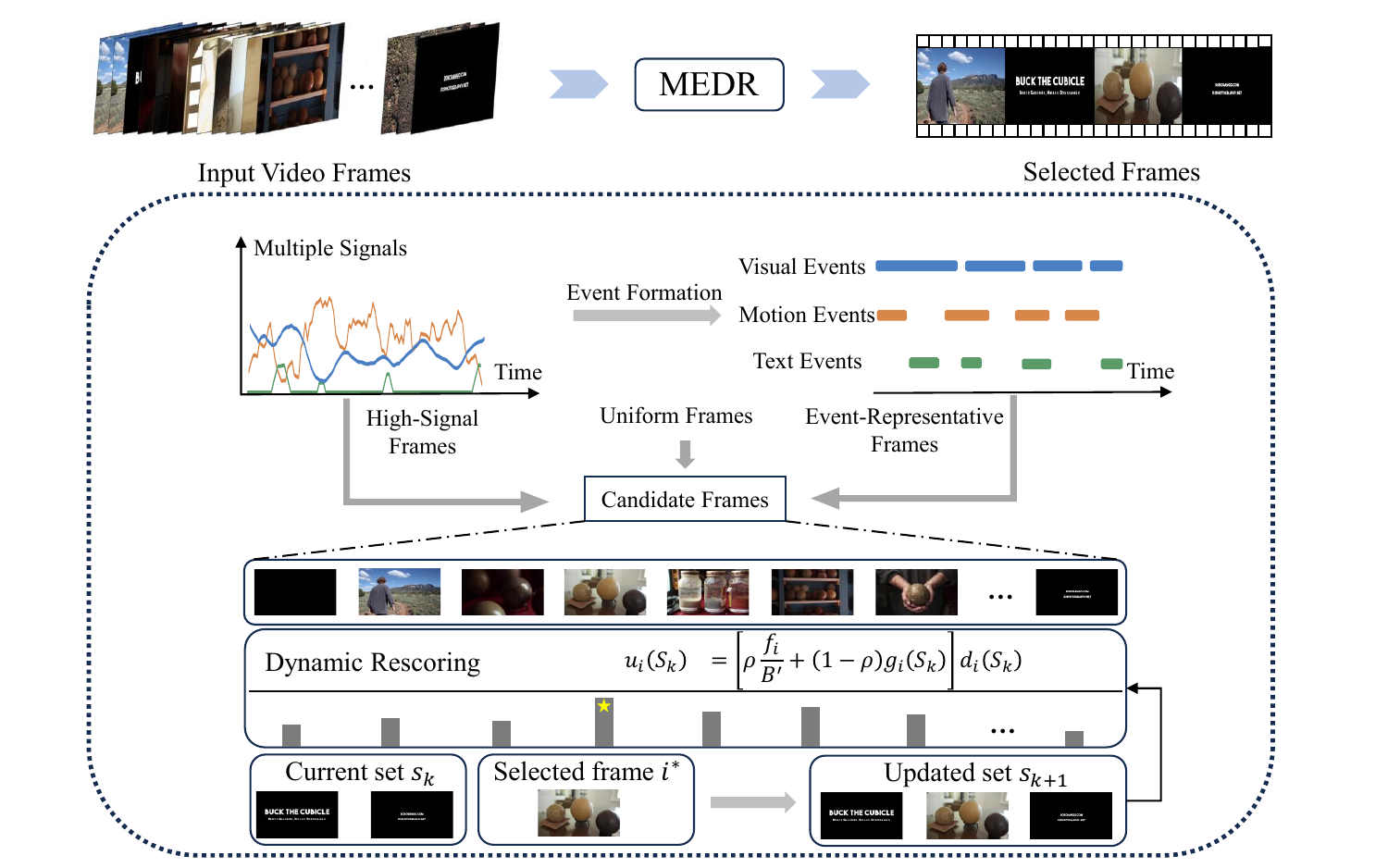}
    \caption{MEDR consists of two stages. Multi-Signal Event Modeling converts
    visual, motion, and text signals into type-specific temporal events and
    frame-level scores, from which a compact candidate set is constructed.
    Dynamic Rescoring evaluates each candidate relative to the current selected
    set and updates the remaining scores after every selection.}
    \label{fig:overview}
\end{figure*}

\subsection{Frame Selection for Video Understanding}

Classical summarization balances representativeness and diversity through
sequential determinantal point processes or learned submodular objectives
\cite{gong2014seqdpp,gygli2015submodular}.

Query-dependent approaches include MLLM-based relevance estimation,
Adaptive Keyframe Sampling, Flexible Frame Selection, and the list-wise
training-free selector MDP$^3$
\cite{hu2025mllmframeselection,tang2025aks,buch2025flexible,sun2025mdp3}.
Pretrained encoders such as CLIP support visual--language matching
\cite{radford2021clip}. Because selections depend on the question,
these methods must repeat retrieval for a new query.

Query-independent methods use only the video. Uniform samples fixed
temporal positions; MaxInfo maximizes the geometric volume of frame
embeddings \cite{li2026maxinfo}. MEDR instead models complementary
temporal events and rescores candidates as event coverage and temporal
distribution evolve.

\section{Method}

\subsection{Problem Setting and Overview}

Let a video be represented by $T$ sampled source frames and their
timestamps,
\begin{equation}
    \mathcal{V}=\{(v_i,t_i)\}_{i=1}^{T}.
\end{equation}
Given a frame budget $B$, we construct a query-independent frame set
\begin{equation}
    \mathcal{S}\subseteq\{1,\ldots,T\},
    \qquad
    |\mathcal{S}|=B'=\min(B,T),
\end{equation}
before any question is observed. The selected frames are restored to
chronological order before being passed to the video MLLM. Since
$\mathcal{S}$ depends only on the video, the same visual input can be
used for different questions about that video.

As shown in Figure~\ref{fig:overview}, MEDR contains two stages.
Multi-Signal Event Modeling organizes visual, motion, and text signals into temporal events and frame-level scores, from which
uniform, event-representative, and high-scoring frames form a
compact candidate set. Dynamic Rescoring then constructs the final
set iteratively by considering each candidate's information,
additional event coverage, and balanced temporal coverage relative to
the current selected set. The remaining candidates are rescored after
every selection.

\subsection{Multi-Signal Event Modeling}

Visual details persist within stable intervals, actions may appear as
short motion bursts, and text can change independently of motion.
MEDR therefore models stability, motion, and text separately.

An event is a consecutive interval detected from one signal and does
not require a semantic action label. We represent it as
\begin{equation}
    e=(\mathcal{I}_e,c_e,s_e),
\end{equation}
where $\mathcal{I}_e$ localizes the signal pattern, $c_e$ provides
a representative candidate frame, and $s_e$ determines the event
priority during candidate construction.

\paragraph{Visual events.}
We compute an adjacent-frame visual-change signal $d_i$ from
color-histogram differences, structural similarity
\cite{wang2004ssim}, and perceptual-hash differences. Strong local
changes separate consecutive visual intervals. After rank normalization
and smoothing, let $\bar d_i$ denote the resulting adjacent-change
signal. The bounded stability channel is
\begin{equation}
    S_i=1-\max(\bar d_{i-1},\bar d_i),
\end{equation}
with zero padding at the two video boundaries. We select the frame with
the largest $S_i$ inside each interval as its representative.

\paragraph{Motion events.}
A motion signal $m_i$ is computed from dense optical-flow
statistics \cite{farneback2003opticalflow}. We use the positive excess of
the high-percentile flow over the mean flow together with its concentration,
so a short burst is emphasized over diffuse camera motion. Consecutive
frames with motion signals above a robust video-specific threshold form a
motion event, represented by the frame with the largest $m_i$.

\paragraph{Text events.}
The OCR signal $O_i$ combines recognition confidence with the appearance or
change of recognized content. Consecutive OCR-bearing frames form one
text event when their normalized text remains similar and their
numerical state is unchanged. A substantial text change,
disappearance, or numerical-state change starts a new event. The
frame with the strongest text signal is used as the representative.

The resulting event bank is
\begin{equation}
    \mathcal{E}
    =\mathcal{E}^{v}\cup\mathcal{E}^{m}\cup\mathcal{E}^{x}.
\end{equation}
Visual-event strength reflects interval length and visual
stability, motion-event strength accumulates motion above
the background level, and text-event strength reflects the
strongest OCR signal in the interval. MEDR directly compares them after applying the same
logarithmic compression:
\begin{equation}
\ell_e = \log(1+\max(s_e,0)).
\end{equation}
The compression reduces the influence of unusually large
raw values, and $\ell_e$ is used as a heuristic event priority
for candidate construction and event weighting.

In addition to events, each frame receives a bounded multi-signal score.
Let $\mathcal{K}\subseteq\{S,A,O\}$ be the enabled channels with support in
the video, where $S_i$, $A_i$, and $O_i$ denote calibrated stability,
concentrated-motion, and OCR evidence. We define
\begin{equation}
    f_i=\frac{1}{|\mathcal{K}|}
    \sum_{k\in\mathcal{K}}k_i,
    \qquad 0\leq k_i\leq 1.
\end{equation}
The denominator counts enabled channels that have support somewhere in
the video. A frame with no evidence in such a channel contributes zero
for that channel; if $\mathcal{K}=\emptyset$, we set $f_i=0$. Removing a
signal omits its channel and the corresponding event family, while the
remaining channels retain their bounded scales.

\subsection{Dynamic Rescoring}

A fixed ranking cannot account for how the value of a candidate
changes after related events or nearby temporal regions have already
been selected. MEDR therefore recomputes candidate utilities after
every set update.

\paragraph{Candidate-frame construction.}
We divide the source timeline into $B'$ equal-duration temporal cells and
first retain one central source frame from every nonempty cell. Let $M$ be
the maximum candidate-set size and
\begin{equation}
    L=\min\bigl(T,\max(M,B')\bigr).
\end{equation}
If $T\leq L$, all source frames are retained. Otherwise, the candidate
set $\mathcal{C}$ is formed by preserving those cell representatives,
then adding $B'$ uniformly spaced frames, event representatives in
descending $\ell_e$, and finally the remaining frames in descending
$f_i$ until $|\mathcal{C}|=L$. The temporal representatives protect
global coverage while the other sources preserve detected events and
strong local signals.

\paragraph{Event--frame affinity.}
An event representative provides a direct candidate for its
interval, but other nearby frames may describe the same event more
clearly. The affinity between event $e$ and candidate $i$ is
therefore computed as
\begin{equation}
    \widetilde{a}_{e,i}
    =
    \tau_{e,i}\phi_{e,i}
    z_i^{\operatorname{type}(e)},
\end{equation}
where temporal compatibility is given by
\begin{equation}
    \tau_{e,i}
    =
    \exp\!\left(
        -\frac{d(t_i,\mathcal{I}_e)}{\sigma_e}
    \right).
\end{equation}
The distance $d(t_i,\mathcal{I}_e)$ is zero inside the event and equals
the distance to its nearest boundary otherwise. The scale $\sigma_e$
is the larger of the event duration and the source sampling interval.

Content compatibility is represented by $\phi_{e,i}$. For visual
and motion events, it measures the maximum lightweight visual
similarity between candidate $i$ and the frames inside
$\mathcal{I}_e$, using perceptual-hash and color-histogram
similarity. For text events, it indicates valid OCR content
associated with the event interval. The type-specific signal
$z_i^{\operatorname{type}(e)}$ is $S_i$, $A_i$, or the
OCR-presence signal for visual, motion, and text events. A high affinity therefore requires temporal
association, compatible content, and a strong value of the
corresponding signal type.

We set $a_{e,i}=\operatorname{clip}(\widetilde{a}_{e,i},0,1)$ and retain events with
positive support. Let $\mathcal{Q}$ be the active event families and
$\mathcal{E}_q$ the active events in family $q$, and
$\mathcal{E}_{\mathrm{act}}=\bigcup_{q\in\mathcal{Q}}\mathcal{E}_q$.
Event priorities give
each active family the same total mass while preserving strength within
that family:
\begin{equation}
    w_e
    =
    \frac{1}{|\mathcal{Q}|}
    \frac{\ell_e}{\sum_{e'\in\mathcal{E}_{q(e)}}\ell_{e'}},
    \qquad e\in\mathcal{E}_{q(e)}.
\end{equation}
This prevents a fragmented OCR stream or a large motion residual from
dominating the other evidence families.

\paragraph{Additional event coverage.}
After $k$ iterations, the selected set $\mathcal{S}_k$ represents
event $e$ to the degree
\begin{equation}
    c_e(\mathcal{S}_k)
    =
    \max_{j\in\mathcal{S}_k}a_{e,j},
    \qquad
    c_e(\emptyset)=0.
\end{equation}
The additional event coverage contributed by an unselected
candidate $i$ is
\begin{equation}
    g_i(\mathcal{S}_k)
    =
    \sum_{e\in\mathcal{E}_{\mathrm{act}}}
    w_e
    \left[
        a_{e,i}-c_e(\mathcal{S}_k)
    \right]_+.
\end{equation}
Only improvements beyond the current representation are rewarded.
A candidate receives a large gain when it covers an absent or
weakly represented event, but little gain when another selected
frame already represents that event equally well.

\paragraph{Balanced temporal coverage.}
Event coverage alone cannot prevent different events from producing
several selected frames in one short temporal region. Let $b(i)$ denote
the temporal cell of candidate $i$, and let $n_q(\mathcal{S}_k)$ be the
number of selected candidates in cell $q$. At each step we first restrict
the eligible candidates to cells with minimum occupancy. Within those
cells, temporal separation is represented by the normalized distance to
the nearest selected timestamp. Using the full sampled-source time span,
\begin{equation}
    D=\max_{1\leq i\leq T}t_i-\min_{1\leq i\leq T}t_i,
    \qquad \Delta=\frac{D}{B'},
\end{equation}
we define the normalized temporal distance as
\begin{equation}
\begin{aligned}
d_i(\mathcal{S}_k)=
\begin{cases}
1, & k=0,\\
\min\!\left(1,
\frac{\min_{j\in\mathcal{S}_k}|t_i-t_j|}
{\Delta}
\right),
& k>0.
\end{cases}
\end{aligned}
\end{equation}
The cell constraint and this linear spacing factor address different
forms of repetition: the former balances global occupancy, while the
latter discourages nearby duplicate selections.
Writing $m_k=\min_{j\in\mathcal{C}\setminus\mathcal{S}_k}
n_{b(j)}(\mathcal{S}_k)$, the eligible set is
\begin{equation}
    \mathcal{C}_k=
    \left\{i\in\mathcal{C}\setminus\mathcal{S}_k:
    n_{b(i)}(\mathcal{S}_k)=m_k\right\}.
\end{equation}

\paragraph{Dynamic score and set update.}
Candidate utility combines frame-level information, additional
event coverage, and temporal separation:
\begin{equation}
    \begin{aligned}
    u_i(\mathcal{S}_k)
    &=\left[\rho\frac{f_i}{B'}+(1-\rho)g_i(\mathcal{S}_k)\right]
    d_i(\mathcal{S}_k).
    \end{aligned}
\end{equation}
The first factor balances the normalized information of an individual
frame against the new event evidence it contributes. The spacing factor
reduces the value of candidates close to the current selected set.

Starting from $\mathcal{S}_0=\emptyset$, the next frame is selected
according to
\begin{equation}
    \begin{aligned}
    i^*&=\arg\max_{i\in\mathcal{C}_k}u_i(\mathcal{S}_k),\\
    \mathcal{S}_{k+1}&=\mathcal{S}_k\cup\{i^*\}.
    \end{aligned}
\end{equation}
until $|\mathcal{S}_k|=B'$. Signals, events, and the event--frame
affinity matrix are computed once for each video, while cell occupancy,
event coverage, and temporal spacing are updated as the selected set
grows. The remaining candidates are consequently rescored after every
iteration.

\section{Experiments}
\label{sec:experiments}
\begin{table*}[t]
\centering
\small
\setlength{\tabcolsep}{3.5pt}
\begin{tabular}{@{}llcc|cc@{}}
\toprule
& & \multicolumn{2}{c|}{Qwen2.5-VL-7B} & \multicolumn{2}{c}{Qwen3-VL-8B} \\
Method & $B$ & V-MME & LongVideoBench & V-MME & LongVideoBench \\
\cmidrule(lr){3-4}\cmidrule(l){5-6}
\midrule
AKS$^{\dagger}$ & 32 & 61.37 & 60.58 & 65.48 & 62.08 \\
AKS$^{\dagger}$ & 64 & 62.30 & 59.99 & 68.48 & 64.17 \\
\midrule
Uniform & 32 & 58.89 & 58.26 & 63.37 & 60.43 \\
Uniform & 64 & 61.85 & 57.44 & 66.44 & 62.38 \\
MaxInfo & 32 & 59.37 & 58.26 & 63.67 & 59.99 \\
MaxInfo & 64 & 61.74 & \textbf{59.16} & 66.41 & 62.23 \\
\midrule
MEDR & 32 & 60.11 & 58.04 & 64.26 & 60.28 \\
MEDR & 64 & \textbf{63.63} & 59.09 & \textbf{67.78} & \textbf{62.45} \\
\bottomrule
\end{tabular}
\caption{Accuracy (\%) on Video-MME (V-MME) and LongVideoBench. Bold values indicate the best matched query-independent result at 64 frames. AKS$^{\dagger}$ is query-dependent. Budgets are upper bounds; AKS can return fewer frames. The evaluation contains 2,700 Video-MME and 1,337 LongVideoBench questions.}
\label{tab:main_results}
\end{table*}

\subsection{Experimental Setup}
\label{sec:experimental-setup}

\paragraph{Datasets and evaluation.}
We evaluate MEDR on Video-MME and LongVideoBench
\cite{fu2025videomme,wu2024longvideobench}. Video-MME contains 2,700
questions over 900 videos, while LongVideoBench contains 1,337 questions
over 753 videos. For MEDR and the query-independent baselines, frames
are selected before the question is observed and the resulting set is
reused for every question associated with that video. Accuracy is computed using the official
multiple-choice evaluation protocol.

\paragraph{Models and baselines.}
We use Qwen2.5-VL-7B-Instruct
\cite{bai2025qwen25vl} and Qwen3-VL-8B-Instruct
\cite{bai2025qwen3vl}. Uniform sampling and MaxInfo
\cite{li2026maxinfo} are included as query-independent reference methods.
The query-independent comparison uses matched source videos, frame
budgets, and backbone-specific inference settings. Adaptive Keyframe
Sampling (AKS) \cite{tang2025aks} is evaluated with the same inference
profiles as a query-dependent reference. Its existing question-conditioned
selections are reused, with up to the requested number of frames.

\paragraph{Implementation details.}
For MEDR, source-frame features are sampled at approximately 1~fps.
If this produces fewer than 128 positions, the source grid is replaced by
uniformly spaced positions over the video, up to the available raw-frame
count. The verified caches contain at least 128 source positions for every
evaluated video. The method selects up to $B$ frames in general and
produces exactly 32 or 64 frames in the corresponding reported runs.
The internal candidate set contains at most 512 frames. MEDR constructs
signal-specific events once per video and dynamically updates utilities
while balancing occupancy across temporal cells. We use $\rho=0.85$ at
both budgets; the selection path uses lightweight visual features,
optical flow, and OCR, without a CLIP encoder.

Uniform samples evenly spaced positions on the raw-video frame axis.
MaxInfo uses its pretrained CLIP candidate features and the requested
output budget. The methods share the final frame budget, while retaining
their respective candidate construction. For each backbone, the matched
query-independent runs use the same prompts, visual preprocessing,
greedy decoding, and seed. Qwen2.5-VL and Qwen3-VL retain their respective
backbone-specific visual preprocessing settings.

\subsection{Main Results}

Table~\ref{tab:main_results} reports the 64-frame and 32-frame
budgets. At 64 frames, MEDR reaches 63.63\% and 67.78\% on Video-MME
with Qwen2.5-VL-7B and Qwen3-VL-8B, respectively. These exceed the
strongest matched query-independent baseline by 1.78 and 1.33 percentage
points. On LongVideoBench, Qwen2.5-VL MEDR achieves 59.09\%, within
one correctly answered question of MaxInfo at 59.16\%. With Qwen3-VL,
MEDR achieves 62.45\%, slightly above Uniform at 62.38\% and MaxInfo
at 62.23\%. The clearest aggregate advantage is therefore on Video-MME.

At 32 frames, MEDR obtains 60.11\% and 64.26\% on Video-MME, and
58.04\% and 60.28\% on LongVideoBench. 
The matched 32-frame Video-MME comparison yields gains of 0.74 and 0.59 percentage points over the strongest query-independent baseline for the respective backbones.
Reducing the frame budget from
64 to 32 lowers Video-MME accuracy by 3.52 percentage points for each
backbone, and LongVideoBench accuracy by 1.05 and 2.17 points,
respectively. Both budgets retain one query-independent frame set per
video that can be reused across its questions.

\subsection{Fixed-Set Multi-Question Evaluation}
\label{sec:multi-turn}

To test whether one selected frame set remains useful when the question
changes, we reuse a single set for the three questions associated with each
Video-MME video. Query-independent methods construct this set without
observing any question and keep it unchanged for every turn.
All methods use the same six
question-order seeds. The first turn contains the shared visual input
and its question. Each later turn reuses the same visual input and includes
all preceding questions and model-generated answers before the new question.
For AKS, we use the frame sequence selected for the first-turn question
throughout the sequence, so it also satisfies the fixed-set protocol. This
design measures cross-question reuse rather than repeated frame selection.

\begin{table}[t]
\centering
\small
\setlength{\tabcolsep}{4pt}
\begin{tabular}{lccc}
\toprule
Method & Overall & First Turn & Follow-up \\
\midrule
Uniform & 66.06$\pm$0.27 & 65.39$\pm$1.35 & 66.40$\pm$0.53 \\
MaxInfo & 66.59$\pm$0.33 & 65.50$\pm$1.22 & 67.14$\pm$0.87 \\
AKS$^{\dagger}$ & 66.66$\pm$0.57 & \textbf{66.87$\pm$1.34 } & 66.56$\pm$0.80 \\
MEDR & \textbf{67.32$\pm$0.40} & 66.43$\pm$1.67 & \textbf{67.77$\pm$0.36} \\
\bottomrule
\end{tabular}
\caption{Fixed-set multi-question accuracy (\%) on Video-MME with Qwen3-VL-8B under a budget of up to 64 frames. Results are mean$\pm$sample standard deviation over six question-order seeds. The same selected frame set is reused across all three questions.}
\label{tab:multiturn}
\end{table}

The same visual frame set is retained throughout each multi-question evaluation sequence.
The first-turn and follow-up columns separate initial answering from
answering with a growing dialogue history. Reported variability comes
from six per-video question-order seeds, jointly processed with identical
visual prefixes.
The calibrated MEDR evaluation follows the same fixed-set protocol and
achieves the best overall and follow-up accuracy among the compared methods.

% The archived cases.jpg mixes old predictions and is not used for current Lite claims.
\subsection{Component-Removal Ablations}
\begin{figure*}[t]
    \centering
    \includegraphics[width=\textwidth]{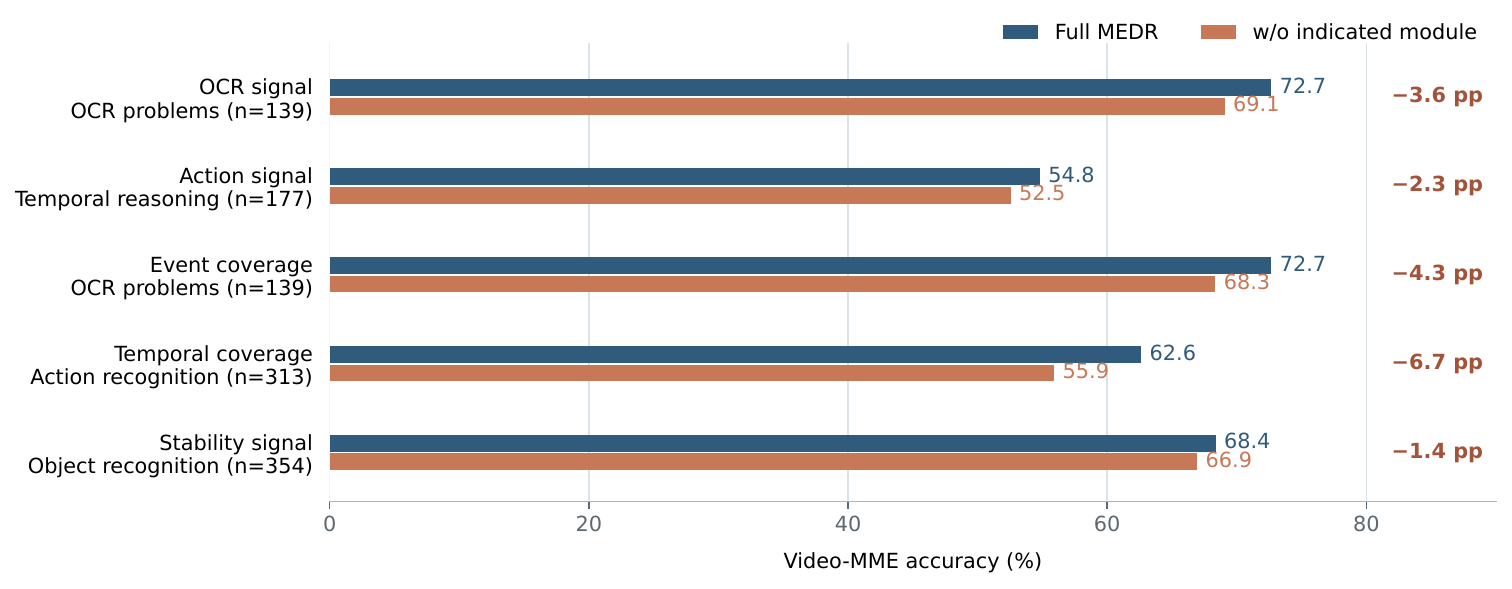}
    \caption{Representative module-to-type effects on Video-MME with
    Qwen2.5-VL-7B. Each pair compares Full MEDR with the method after
    removing the indicated module. Annotations give the accuracy
    decrease in percentage points, and $n$ is the number of questions of
    that type. These selected illustrative effects complement the
    aggregate ablation in Table~\ref{tab:ablation}.}
    \label{fig:component-type-drops}
\end{figure*}

\begin{table}[t]
\centering
\small
\setlength{\tabcolsep}{6.5pt}
\begin{tabular}{@{}lcc@{}}
\toprule
Variant & Video-MME & LongVideoBench \\
\midrule
Full MEDR & \textbf{63.63} & \textbf{59.09} \\
w/o stability signal & 62.93 & 58.49 \\
w/o action signal & 62.33 & 58.71 \\
w/o OCR signal & 61.89 & 57.74 \\
w/o event coverage & 63.26 & 58.34 \\
w/o temporal coverage & 61.44 & 56.39 \\
\bottomrule
\end{tabular}
\caption{Component-removal ablations with Qwen2.5-VL-7B at 64 frames. Signal removals also omit the corresponding event family and rebuild candidates. The other two variants disable event-coverage scoring or both temporal-cell balancing and spacing. Backbone and inference settings are fixed. Values are accuracy (\%) over all 2,700 Video-MME and 1,337 LongVideoBench questions.}
\label{tab:ablation}
\end{table}

Table~\ref{tab:ablation} isolates the contribution of each module with
Qwen2.5-VL-7B. Removing balanced temporal coverage causes the largest degradation on
both datasets, supporting the value of balancing evidence across the
video rather than relying on event coverage alone. Removing OCR also produces a
clear decrease, especially on Video-MME, indicating that text changes supply
evidence that is not reliably recovered from appearance and motion. The
stability and action signals provide complementary improvements, while
event coverage supplies an additional gain by rewarding events that are
not yet represented. The visual similarity term remains part of the
event--frame affinity used by the complete method, but it is treated as an
internal implementation detail rather than a separate top-level component.

To connect these aggregate changes to the intended evidence types, Table~\ref{tab:component-reading}
summarizes the aggregate Qwen2.5-VL Video-MME decreases and links each
removed component to one illustrative question type in
Figure~\ref{fig:component-type-drops}. The type-level effects are
computed over all questions carrying the corresponding benchmark label;
their magnitudes need not follow the aggregate ranking. These associations
illustrate complementary evidence requirements rather than a task-type
tuning rule.

\begin{table}[t]
\centering
\small
\setlength{\tabcolsep}{4pt}
\begin{tabular}{@{}lrl@{}}
\toprule
Removed component & $\Delta$ V-MME & \shortstack[l]{Illustrative\\question type} \\
\midrule
OCR signal & $-1.74$ & OCR problems \\
Action signal & $-1.30$ & temporal reasoning \\
Event coverage & $-0.37$ & OCR problems \\
Temporal coverage & $-2.19$ & action recognition \\
Stability signal & $-0.70$ & object recognition \\
\bottomrule
\end{tabular}
\caption{Selected component effects on Qwen2.5-VL Video-MME. Each change is the accuracy difference from Full MEDR; negative values indicate degradation after removing the component.}
\label{tab:component-reading}
\end{table}

Figure~\ref{fig:component-type-drops} exposes information that is hidden by
the aggregate scores in Table~\ref{tab:ablation}. Each row pairs a removed
module with an evidence-aligned Video-MME question type and shows the full
method against the corresponding ablation. The selected examples all show a
drop after removal. For example, temporal-reasoning accuracy decreases
from 54.80\% to 52.54\% without the action signal, while
action-recognition accuracy decreases from 62.62\% to 55.91\% without
temporal coverage. Removing OCR lowers OCR-question accuracy from
72.66\% to 69.06\%; removing event coverage lowers the same type to
68.35\%. These are selected type-level effects from the same verified
Lite runs used in Table~\ref{tab:ablation}.

\subsection{Qualitative Analysis}

\begin{figure*}[t]
\centering
\includegraphics[width=\textwidth]{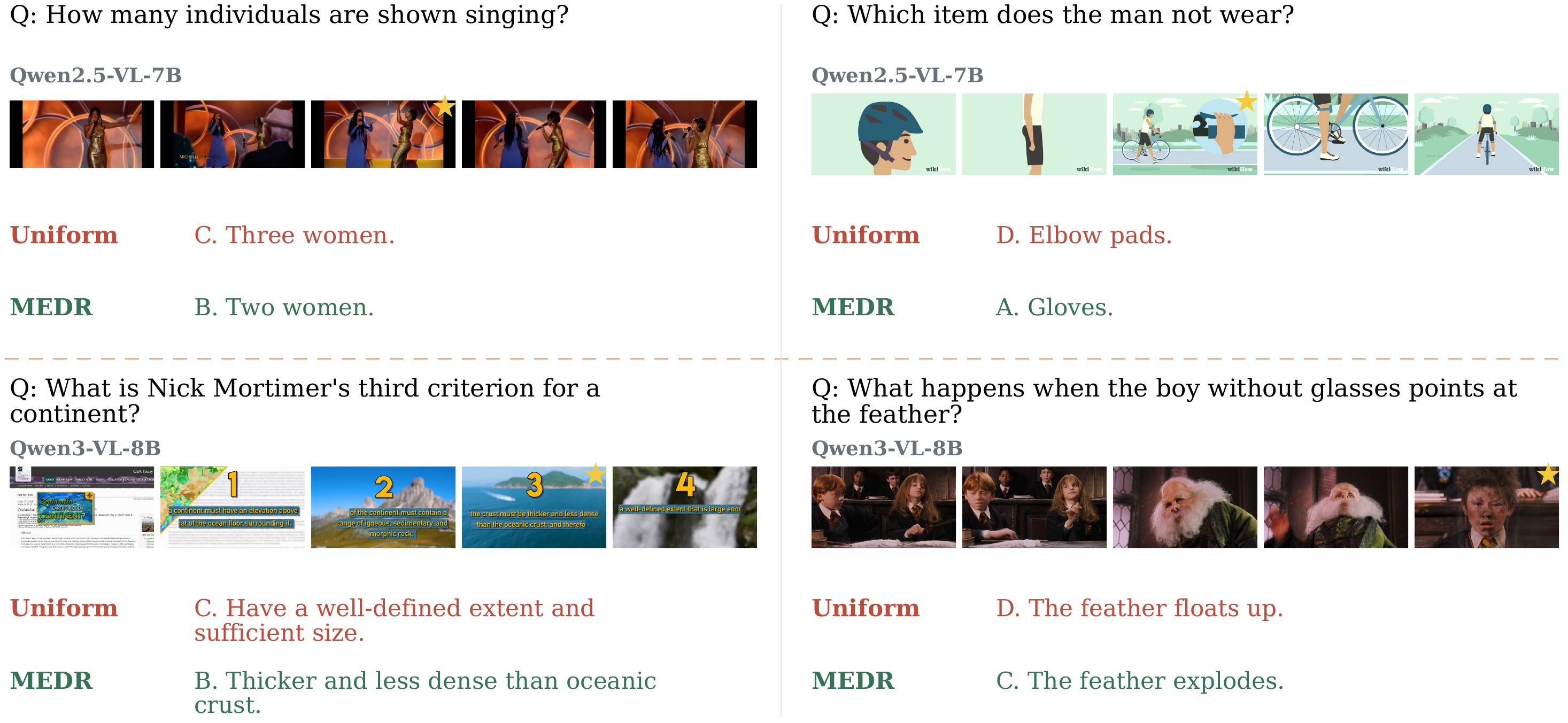}
\caption{Selected Video-MME cases from MEDR and
matched Uniform evaluations with a 64-frame budget. Each panel shows
five representative frames from MEDR's actual selected set, the backbone,
and the two verified predictions. Green answers match the reference;
red answers do not. Stars mark illustrative visual evidence. Questions
are abbreviated; the backbone is shown above each panel.}
\label{fig:qualitative-cases}
\end{figure*}
\label{sec:qualitative-analysis}

Figure~\ref{fig:qualitative-cases} gives traceable examples from the
matched 64-frame Video-MME runs. The examples are selected cases where
Uniform answers incorrectly and MEDR answers correctly with
the same backbone. The counting example shows two singers together, while
the equipment example includes a close view of the cyclist's bare hand.
The text-reading example distinguishes the third continent criterion
from the fourth, and the action example includes the aftermath of the
feather spell. These selected successes illustrate evidence present in
the current frame sets; they do not isolate the contribution of any
individual module.

\section{Conclusion}
MEDR is a training-free and query-independent frame selection method that
combines Multi-Signal Event Modeling with Dynamic Rescoring. It organizes
visual, motion, and text signals into signal-specific temporal events
and updates candidate scores as the selected set grows, producing a fixed visual
input that can be reused across questions. Matched experiments show
64-frame Video-MME gains across two video MLLMs, competitive
LongVideoBench accuracy, and degradation when individual
signals or set-level selection modules are removed. Future work may explore duration-adaptive source-frame
sampling to better preserve brief events while maintaining
the efficiency of frame selection.

\clearpage
\section*{Limitations}
MEDR constructs a query-independent frame set from visual, motion, and OCR
signals. These signals are proxies for useful evidence: visual changes or
persistent text can be unrelated to a question, while small or blurred text
may not be recognized reliably. Event modeling and dynamic rescoring cannot
recover evidence absent from the extracted signals. Moreover, a fixed frame
set cannot adapt to a later question that requires details outside the selected
frames. Signal extraction also incurs preprocessing cost, although the
resulting features and selections can be reused across questions.

Our source-grid construction uses approximately 1~fps with a
uniformly spaced fallback when fewer than 128 positions are obtained.
MEDR selects up to 64 frames in the evaluated settings; the verified
runs contain exactly the requested 32 or 64 frames. Brief events between
sampled timestamps may still be missed. Uniform and MaxInfo retain their
own candidate construction under the same final frame budget. The present evaluation
does not establish the best sampling rate, minimum candidate count, or frame
budget for different video durations.

Our multi-question evaluation uses three questions per Video-MME video with
varied question orderings. It evaluates frame-set reuse in a controlled setting;
it does not cover open-ended conversations with user-generated follow-up
questions or substantially longer dialogue histories. 

\bibliography{aaai2027_30refs_verified}

\end{document}